\documentclass{article}

\usepackage{PRIMEarxiv}

\usepackage[utf8]{inputenc} 
\usepackage[T1]{fontenc}    
\usepackage{hyperref}       
\usepackage{url}            
\usepackage{booktabs}       
\usepackage{amsfonts}       
\usepackage{nicefrac}       
\usepackage{microtype}      
\usepackage{fancyhdr}       
\usepackage{graphicx}       
\graphicspath{{media/}}     

\usepackage[table,svgnames,dvipsnames]{xcolor}         
\usepackage[outline]{contour}
\usepackage{wrapfig}
\usepackage{bm}
\usepackage{bbm}
\usepackage{multirow}
\usepackage{pifont}
\usepackage{rotating}
\usepackage{tikz}
\newcommand{\xmark}{\ding{55}}%

\hypersetup{
	colorlinks=true, 
	linktoc=all,     
	linkcolor=black,  
	citecolor=cyan,
}

\usepackage{physics}

\newcommand{\rr}{\mathbb{R}}

\newcommand{\llangle}{\langle \langle}
\newcommand{\rrangle}{\rangle \rangle}

\DeclareMathOperator*{\argmin}{arg\,min}

\title{Discovering Interpretable Physical Models using Symbolic Regression and Discrete Exterior Calculus}

\author{
	Simone Manti, Alessandro Lucantonio \\
	Department of Mechanical and Production Engineering \\
	Aarhus University\\
	Aarhus, DK\\
	\texttt{\{smanti,a.lucantonio\}@mpe.au.dk}}

\begin{document}
	\maketitle
	
	
	
	\begin{abstract}
		Computational modeling is a key resource to gather insight into physical systems in modern scientific research and engineering. While access to large amount of data has fueled the use of Machine Learning (ML) to recover physical models from experiments and increase the accuracy of physical simulations, purely data-driven models have limited generalization and interpretability. To overcome these limitations, we propose a framework that combines Symbolic Regression (SR) and Discrete Exterior Calculus (DEC) for the automated discovery of physical models starting from experimental data. Since these models consist of mathematical expressions, they are interpretable and amenable to analysis, and the use of a natural, general-purpose discrete mathematical language for physics favors generalization with limited input data. Importantly, DEC provides building blocks for the discrete analogue of field theories, which are beyond the state-of-the-art applications of SR to physical problems. Further, we show that DEC allows to implement a strongly-typed  SR procedure that guarantees the mathematical consistency of the recovered models and reduces the search space of symbolic expressions. Finally, we prove the effectiveness of our methodology by re-discovering three models of Continuum Physics from synthetic experimental data: Poisson equation, the Euler's Elastica and the equations of Linear Elasticity. Thanks to their general-purpose nature, the methods developed in this paper may be applied to diverse contexts of physical modeling.
	\end{abstract}
	
	\section{Introduction}
	
	Science and engineering routinely use computer simulations to study physical systems and gather insight into their functioning. These simulations are based on established models made of sets of differential and algebraic equations that encode some first principles (\textsl{i.e.}~balances of forces, mass, energy\dots), and whose analytical solution cannot be generally found. Given the recent advances in Machine Learning, methods that leverage experimental data to automatically discover models from observations offer a valuable support to scientific research and engineering, whenever the analysis of a system with complex or unknown governing laws is involved. 
	
	To address the question of interpretability, \textsl{Symbolic Regression} (SR) has been employed in several works to discover the mathematical expressions of the governing equations of single-DOF or dynamical systems \cite{schmidt2009distilling, bongard2007automated, brunton2016discovering}. In \cite{udrescu2020ai}, a combination of neural networks and physics-inspired techniques enhance SR by cleverly reducing the search space, while in \cite{cranmer2020discovering} SR was used to distill equations from a Graph Neural Networks with a suitable architecture for problems involving Newtonian and Hamiltonian mechanics. Recent works \cite{petersen2019deep, biggio2021neural} proposed to combine Deep Learning with SR via Recurrent Neural Networks or Transformer Networks. In \cite{rudy2017data, kaptanoglu2021pysindy}, a sparse regression approach was introduced to recover the governing equations of nonlinear dynamical systems and continuum systems governed by PDEs. However, this method produces coordinate-based descriptions of the PDEs that obscure their meaning, it is sensitive to noisy data and is based on a special representation of the equations (linear combination of terms containing the derivatives of the unknown). Apart from \cite{kaptanoglu2021pysindy}, all these papers apply SR to the (re)discovery of algebraic or ordinary differential equations of classical Physics or other standard benchmarks, and are hence inappropriate for field problems, which are a large class of problems in Continuum Physics (fluid/solid mechanics, heat transfer, mass transport, \dots). Moreover, all these approaches do not exploit the geometrical character of Physics, so that, for example, physical variables and mathematical operators may be combined arbitrarily without accounting for their spatial localization (\textsl{i.e.}~whether they are defined on points, lines, surfaces or volumes). A high-level comparison between our framework and existing ones is reported in Table~\ref{tab:comparison}.
	
	Here, we devise a framework that combines SR and \textsl{Discrete Exterior Calculus} (DEC) \cite{grinspun2006discrete,hirani2003discrete} for the discovery of physical models starting from experimental data. Following the pioneering ideas of Tonti \cite{tonti2013mathematical, tonti2014starting} and others \cite{milicchio2008codimension,dicarlo2009chain,dicarlo2009discrete} we adopt a discrete \textsl{geometrical} description of Physics that offers a general-purpose language for physical theories, including the discrete analogue of field theories.  A similar idea was proposed in \cite{behandish2022ai}, although with a less complete mathematical framework than DEC, a different model search strategy was employed and a no application to vector-valued problems was shown. In Section~\ref{sec:learning}, we introduce the main concepts of DEC and we extend the theory to deal with problems characterized by vector-valued unknowns, such as the equations of Linear Elasticity. Then, we describe our approach to SR that exploits the synergy between DEC and SR for the effective identification of mathematically and physically consistent models. In Section~\ref{sec:experiments} we demonstrate that our method can successfully recover the coordinate-independent representations of three models of Continuum Physics starting from a limited set of noisy, synthetic experimental data.
	
	\section{Automated Discovery of Physical Models}\label{sec:learning}
	
	\begin{table}[tb!]
		\caption{Comparison among SR frameworks for physical model identification. Some instances of \xmark \, are software and/or conceptual limitation. We denoted by \xmark$^*$ some limitations of our framework that we plan to overcome next.}
		\begin{center}
			\resizebox{\columnwidth}{!}{
				\begin{tabular}{l l c c c c c c c}
					\toprule
					& & Ours & PySINDy \cite{kaptanoglu2021pysindy, rudy2017data} & EQL \cite{sahoo2018learning} & Eureqa \cite{schmidt2009distilling} & DSR \cite{petersen2019deep} & AI Feynman \cite{udrescu2020ai} & PySR \cite{cranmer2023interpretable}\\
					\midrule
					\rowcolor{Gainsboro!60}
					Field Problems & & \checkmark & \checkmark & \xmark & \xmark & \xmark & \xmark & \xmark \\
					\hfill& Domain source & \checkmark & \xmark& \xmark & \xmark & \xmark &  \xmark &  \xmark\\
					\rowcolor{Gainsboro!60}
					&Stationary/Non-stationary & \checkmark/\xmark$^*$ &  \xmark/\checkmark & \xmark/\xmark & \xmark/\xmark & \xmark/\xmark &  \xmark/\xmark &  \xmark/\xmark \\
					&Variational/Non-variational & \checkmark/\xmark$^*$ & \xmark/\checkmark & \xmark/\xmark & \xmark/\xmark & \xmark/\xmark &  \xmark/\xmark &  \xmark/\xmark \\
					\midrule
					\rowcolor{Gainsboro!60}
					Dynamical systems &  & \xmark &  \checkmark & \checkmark & \checkmark & \xmark &  \xmark &  \xmark \\
					\midrule
					Algebraic equations &  & \checkmark &  \checkmark & \checkmark & \checkmark & \checkmark &  \checkmark &  \checkmark \\
					\bottomrule
			\end{tabular}}
		\end{center}
		\label{tab:comparison}
	\end{table}
	
	As anticipated in the Introduction, we combine SR and DEC to produce interpretable physical models starting from data. In particular, geometry is at the foundations of many physical theories of the macrocosm, \textsl{e.g.} general relativity, electromagnetism, solid and fluid mechanics \cite{tonti2013mathematical}, and DEC allows to capture the geometrical character of these theories better than standard vector calculus \cite{grinspun2006discrete,dicarlo2009discrete, dicarlo2009chain, milicchio2008codimension}. The incorporation of such a mathematical language within SR has several key advantages: 
	\begin{itemize}
		\item being DEC a discrete theory, it bypasses the differential formulation, so that the governing equations can be readily implemented in computations, without the need for discretization schemes;
		\item it allows to distinguish physical variables based on their spatial localization and thus guarantees that the generated symbolic expressions are mathematically and physically consistent; 
		\item it helps to constrain the search space of the SR by suggesting a type system for the mathematical operators and physical variables;
		\item it encodes the discrete counterpart of several differential operators (such as divergence, gradient and curl) in a minimal set of operators, thus making the symbolic representations more compact and expressive.
	\end{itemize}
	We argue that our model discovery strategy exploits these features of DEC to produce compact and effective models that generalize well starting from small training datasets. Also, our approach aims at identifing the governing equations that generate the observed data, rather then directly fitting the observations through an end-to-end model that relates inputs (sources, parameters) to measured quantities. This is a substantial difference with respect to all the state-of-the-art alternatives reported in Table~\ref{tab:comparison}, which is also expected to provide better generalization and physical insight associated to the recovered model. 
	
\begin{figure}[t!]
		\centering
		\includegraphics[scale=0.7]{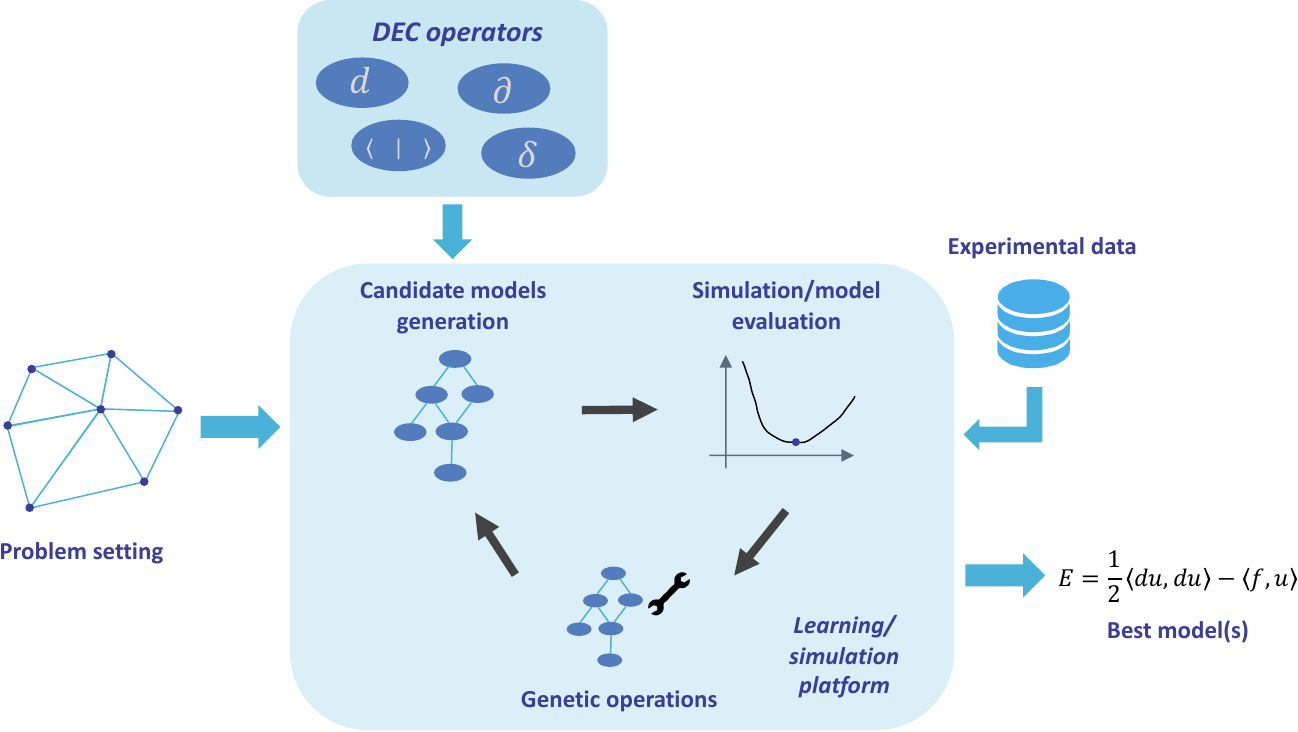}
		\caption{Schematic of the model discovery approach.} 
		\label{fig:machinery}
	\end{figure}

	A schematic of the proposed framework is depicted in Figure~\ref{fig:machinery}. Model discovery consists of several steps. In the problem setting step, the physical system whose model has to be identified is defined in terms of a discrete structure, i.e. a ``mesh'', resembling (a simplified/coarse-grained description of) its geometry. A preliminary dimensional analysis is carried out to define the appropriate scales for the physical quantities and the number of dimensionless groups, including those that will be used as fitting parameters as a part of SR. Physical quantities, including those that are measured experimentally, are defined in terms of their spatial localization and adimensionalized. The set of DEC operators and other mathematical operations to be used for SR is chosen. The definition of boundary conditions and domain sources (such as external forces) is also carried out at this stage. Then, SR is performed based on the following loop:
	\begin{enumerate}
		\item candidate models generation: a population of trees representing mathematical models based on the chosen operators set is generated, initially in a random fashion; 
		\item each model is solved, \textsl{i.e.}~by interpreting it as the expression of a potential energy function and minimizing it, and the solution (or a derived physical quantity) is compared with experimental data to evaluate an error metric that contributes to the fitness of each model; 
		\item the candidate models are evolved through \textsl{genetic operations} (crossover and mutation) based on their fitness values and produce a new generation. 
	\end{enumerate} 
	The loop repeats usually until reaching a given number of generations, after which the top performing model(s) can be extracted.
	
	In the following sections, we first briefly review some fundamental concepts of DEC, and then describe the SR approach based on it, with particular reference to physical systems that can be described by a potential energy (variational formulation).
	
	\subsection{Elements of Discrete Exterior Calculus (DEC)}
	\label{DiscreteCalculus}
	\textsl{Exterior Calculus} (EC) deals with objects called differential forms and their manipulations (differentiation, integration) over smooth manifolds. Compared to classical tensor calculus using index notation, it provides a coordinate-independent description of physical models and unifies several differential operators, which results in more compact and readable equations that easily generalize to manifolds of different dimension \cite{grinspun2006discrete}. 
	DEC offers an \textsl{inherently discrete} counterpart to EC, designed to preserve some fundamental differential properties. In particular, in the context of DEC, \textsl{cell-complexes} and \textsl{cochains} play the role of smooth manifolds and differential forms, respectively. 
	Here, we provide a very brief introduction to DEC, geared towards the applications we will consider in Section~\ref{sec:experiments}.
	We report in Appendix \ref{DEC details} more formalism of DEC definitions and notations.
	Without loss of generality, we limit the presentation to simplicial complexes.  A more detailed treatment of DEC can be found in \cite{grinspun2006discrete,hirani2003discrete, grady2010discrete}.
	
	Henceforth, $\mathcal{E}^N$ is the embedding $N$-dimensional Euclidean point space, $\mathcal{V}$ is the associated space of translations associated with $\mathcal{E}^N$ and $\text{Lin}$ is the space of linear applications from $\mathcal{V}$ to $\mathcal{V}$ (tensors).

	From the physical modeling viewpoint, a simplicial complex provides the topological and geometrical structure that supports the physical phenomenon. 
	A generic simplex is indicated with the letter $\sigma$ and we use a pair of subscripts to identify each of them, such that $\sigma_{p,i}$ is the $i$-th $p$-simplex. Cochains are the discrete counterpart of differential forms, hence they are also known as \textsl{discrete forms}. Specifically, a
	cochain can describe a physical quantity (differential form) integrated on the corresponding simplex and is thus the proxy for a \textsl{field}.
	The discrete analogue of the exterior derivative in Exterior Calculus is represented by the \textsl{coboundary} $d$, which plays a key role in expressing \textsl{balance statements} \cite{tonti2013mathematical} of physical models.
	A \textsl{dual complex} is needed to represent the behavior of a physical quantity in the neighborhood of a simplex (see Section 4.2 in \cite{tonti2013mathematical}); a role similar to that of the dual complex is played by, \textsl{e.g.}, stencils in the finite difference method.
	
	In addition to the topological operators introduced so far, metric-dependent ingredients are needed to complete the formulation of physical theories, especially as concerns constitutive equations. Indeed, these equations relate primal to dual variables \cite{tonti2013mathematical, milicchio2008codimension}, so that it is convenient to introduce an isomorphism between primal and dual cochains, called the \textsl{discrete Hodge star} and denoted by $\star$.
	Finally, in continuum theories the \textsl{$L^2$ inner product} allows to express power expenditures by pairing power-conjugated quantities and thus allows to represent \textsl{potential energy terms} in variational models for physical systems, which will be employed in our numerical experiments. In particular, it acts on a pair of primal/dual cochains and in the following is indicated by $\langle , \rangle$. This inner product allows to define the \textsl{discrete codifferential} $\delta$ as the adjoint of the coboundary with respect to this operator.

	\subsection{Symbolic regression}
	\label{sec:symbreg}
	We tackle SR through Genetic Programming \cite{koza1994genetic} (GP), an evolutionary technique that explores the space of symbolic models starting from an initial population and using genetic operations. The candidate models (\textsl{individuals}) are represented by expression trees, whose nodes can be either \textsl{primitives} (\textsl{i.e.}~functions and operators from a prescribed pool) or \textsl{terminals} (\textsl{i.e.}~variables and constants).
	Specifically, for the primitives we use the standard operations on scalars and the DEC operations introduced in Section~\ref{DiscreteCalculus}, see Table~\ref{tab:primitives}.

	The initial population is generated using the ramped half-and-half method (minimum height $=2$, maximum height $=5$) that allows to obtain trees with a variety of sizes \cite{o2009riccardo}. We constrain the initial individuals to have length -- the number of nodes of its tree -- less or equal than $100$ and to contain the unknowns of the problem.
	
	\paragraph{Model discovery strategy} The goal of the GP evolution is to \textsl{maximize} the following fitness measure
	\begin{align}
		\label{eq:fitness}
		F(I) := -(\alpha\,\mbox{MSE}(I) + \eta\,R(I)),
	\end{align}
	where $\alpha$ is a scaling factor set equal to $\{1, 10, 1000\}$ for \textsl{Poisson}, \textsl{Elastica} and \textsl{Linear Elasticity}, respectively, MSE is the mean-square error of the solution corresponding to the current individual $I$ and $R$ is a regularization term, and $\eta$ is the associated hyperparameter. The regularization term aims at preventing overfitting, \textsl{i.e.}~high fitness on the training dataset and low fitness on the validation dataset. To favor individuals with low complexity among those who accurately fit the training dataset, we represent the regularization term as $R(I) := \text{length}(I)$.
	
	Starting from the initial population, the program evolves such a population through the operations of \textsl{crossover} and \textsl{mutation}, similarly to genetic algorithms. Individuals eligible for crossover and mutation are chosen through a binary \textsl{stochastic tournament} selection, where the competing individuals have given survival probabilities such that the weakest individual too has a chance of winning the tournament.
	In all the experiments we used one-point crossover and a \textsl{mixed mutation} operation that is chosen among uniform mutation, node replacement and shrink (see \href{https://deap.readthedocs.io/en/master/}{here} for details on these operations). The two survival probabilities for the stochastic tournament and the three probabilties associated to the mixed mutation are hyperparameters tuned as a part of the model selection.
	
	As anticipated, the discrete mathematical framework introduced in the previous section naturally suggests a type system for the primitives of the GP procedure. For example, we cannot sum cochains with different dimensions. To enforce this typing, we use a variant of Genetic Programming called \textsl{Strongly Typed Genetic Programming} (STGP), which generates and manipulates trees there are type-consistent. Crucially, in our approach the physical consistency of the generated expressions is guaranteed by the use of \textsl{dimensionless quantities}. To evolve the population, we use a $(\mu + \lambda)$-{Genetic Algorithm} with $\mu = \lambda$ and overlapping generations, such that top $\mu$ individuals with the highest fitness are selected (truncation selection) among the total population including parents and offspring \cite{de2016evolutionary}.

	\paragraph{Datasets}  For each physical system analyzed in our experiments (Section~\ref{sec:experiments}), we generate a dataset based either on exact solution of the discrete model that we would like to recover or on the numerical solution of the corresponding continuous problem, for different values of a chosen load parameter. We split the dataset in three sets (\textsl{double hold out}): training  $(50\%)$, validation  $(25\%)$, and test  $(25\%)$. In the present context, training consists of model discovery and possibly fitting of dimensionless parameters with unknown values. By a screening phase we identify the hyperparameters that mostly affect the learning process and we tune their values through a \textsl{grid search} to minimize the MSE on the validation set. More details on the grid search choices and results are reported in Appendix \ref{sec:gridsearch}. With the best values resulting from the grid search (Table~\ref{tab:hyperparams}), we perform 50 model discovery runs using as a dataset the combination of the training and the validation sets. Finally, we assess the generalization capability of the best model found in the model discovery runs by evaluating its MSE on the test set.

	\paragraph{Variational models of physical systems}	We  model physical systems adopting a variational approach, such that each individual of the population that is evolved via SR corresponds to a \textsl{discrete potential energy functional}. A typical individual tree is represented in Figure~\ref{fig:poisson_final}B. Boundary conditions are enforced by adding a penalty term to the discrete energy. 
	For the fitness evaluation, we first compute the minimum-energy solution of each model numerically.  We assign the arbitrary value $10^5$ to the MSE of constant energies or for which the minimization procedure does not converge. The minimum-energy solutions are then compared with those in the datasets to compute the fitness of the individuals according to eq.~\eqref{eq:fitness}.
	
	\subsection{Implementation}
	We implement our framework in two open-source Python libraries: \href{https://github.com/alucantonio/dctkit}{\texttt{dctkit}} (Discrete Calculus Toolkit) for the DEC concepts described in Section~\ref{DiscreteCalculus} and \href{https://github.com/alucantonio/alpine}{\texttt{alpine}} for the symbolic regression. 
	Specifically, \texttt{dctkit} leverages the \texttt{JAX} \cite{jax2018github} library as a high-performance numerical backend for the manipulation of arrays and matrices encoding cochains and DEC operators. For the minimization of the discrete energies found during the symbolic regression process, we use the LBFGS algorithm implemented in the \texttt{pygmo} \cite{Biscani2020} library, with the gradients evaluated via the automatic differentiation feature of \texttt{JAX}. The \texttt{alpine} library is based on the STGP algorithms provided by a custom version of the \texttt{DEAP} (Distributed Evolutionary Algorithms in Python) \cite{DEAP_JMLR2012} library and on the distributed computing features of the \texttt{ray} library \cite{ray}.
	
	\begin{table}[t!]
		\caption{Values of the hyperparameters used in the numerical experiments of Section~\ref{sec:experiments}. For mixed mutation, the probabilities refer to uniform mutation, node replacement and shrink, respectively. For stochastic tournament, the probabilities refer to the survival change of the strongest and the weakest individual, respectively.}
		\begin{center}
				\begin{tabular}{l c c c}
					\toprule
					Hyperparameter & \multicolumn{3}{c}{Value}\\
					& \textsl{Poisson} & \textsl{Elastica} & \textsl{Linear Elasticity}\\
					\midrule
					Number of individuals ($\mu$)  & 2000 & 2000 & 2000 \\
					Crossover/mutation probabilities & (0.2, 0.8) &  (0, 1) & (0.2,0.8)\\
					Mixed mutation probabilities &  (0.8, 0.2, 0) & (0.8, 0.2, 0) & (0.7, 0.2, 0.1)\\
					Stochastic tournament probabilities  &  (0.7, 0.3) & (1, 0) & (0.7,0.3)\\
					Regularization factor ($\eta$)  & 0.1 & 0.01 & 0\\
					\bottomrule
				\end{tabular}
		\end{center}
		\label{tab:hyperparams}
	\end{table}

	\section{Results}
	\label{sec:experiments}
	In this section, we assess the effectiveness of our model discovery method by solving three problems:  \textsl{Poisson} \cite{podio2019continuum},  \textsl{Euler's Elastica} \cite{audoly2000elasticity} and  \textsl{Linear Elasticity}  \cite{gurtin1982introduction}. 
	In the \textsl{Poisson} and \textsl{Linear Elasticity} problems, the goal is to recover the governing equations of the discrete model that is used to generate the dataset. In particular, by running the SR algorithm multiple times, these benchmarks allow to evaluate the recovery rate, \textsl{i.e.}~percentage of runs where the exact equations of the model used to generate the data were found. Instead, in \textsl{Elastica} we emulate a real scenario where experimental data is employed for model discovery. 
	
	The primitives used for the problems are reported in Table~\ref{tab:primitives}. Note that for \textsl{Elastica} we also use $\mathbbm{1}$ as the identity dual $0$-cochain and $\mathbbm{1}_{\text{int}}$ as a primal $0$-cochain that is identically $1$ on the interior of the simplicial complex and $0$ on the boundary nodes\footnote{This primitive is needed to sample the \textsl{curvature} of the rod axis at the interior nodes of the primal complex, since it is not well-defined at the boundary nodes. Curvature is a fundamental ingredient of the elastic energy of a rod \cite{audoly2000elasticity}.}. 
	For all the problems, in addition to the primitives we employ the constant terminals $\{1/2, 2, -1\}$; for \textsl{Linear Elasticity} we also add the constants $\{10, 0.1\}$. 
	
	\subsection{Poisson equation in 2D} \label{sec:Poisson}
	
	In the continuous setting, the Poisson equation can be obtained as the stationarity conditions of the Dirichlet functional \cite{evans2010partial}. Such a functional contains the $L^2$ norm of the gradient of the unknown field. In the discrete setting, we represent the dimensionless unknown as a primal scalar-valued $0$-cochain $u$, and application of the coboundary operator $d$ leads to a primal scalar-valued $1$-cochain that is the discrete equivalent of the gradient of the continuous field integrated along the edges. Denoting by $f$ the  primal scalar-valued $0$-cochain representing the (dimensionless) source, we can write the potential energy function as 
	\begin{align}
		\mathcal{E}_{\text{p}}(u) := \frac{1}{2} \langle du, du \rangle - \langle u, f \rangle = \frac{1}{2} \langle \delta du, u \rangle - \langle u, f \rangle. \label{eq:discDirichlet}
	\end{align}
	It can be shown \cite{desbrun2005discrete} that the stationarity conditions of $\mathcal{E}_{\text{p}}$ correspond to the discrete Poisson equation. The benchmark consists in recovering eq.~\eqref{eq:discDirichlet} defined over a 2-simplicial complex. In particular,  a Delaunay triangulation of the unit square with $230$ nodes was  constructed using \texttt{gmsh} \cite{geuzaine2009gmsh}.

	\begin{figure}[tb!]%
		\centering
		\includegraphics[scale=1]{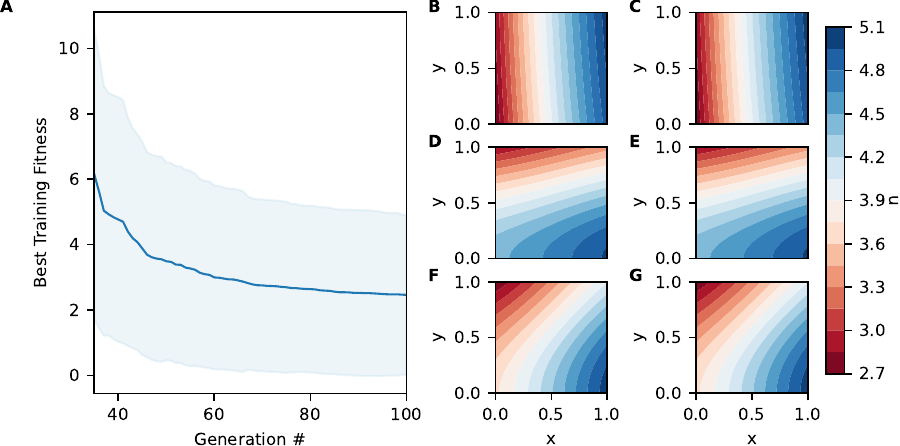}
		\caption{Results for the \textsl{Poisson} problem. (A) Evolution of the mean and standard deviation of the absolute value of the best training fitness (\textsl{i.e.}~fitness of the best individual of a generation evaluated on the training set) over the model discovery runs $(n=50)$ (the first 34 generations have been removed from the plot, since the range of values of the fitness is very large). Comparison between the minimum-energy solutions (B, D, F) corresponding to the best individual at the end of the model discovery stage evaluated for the source cochain associated to the test data $\{u_{2,3}, u_{1,3}, u_{1,0}\}$ of eq.~\eqref{eq:ui} (C, E, G).}
		\label{fig:poisson_plots}%
	\end{figure}
	
	For the SR procedure, we generated $12$ arrays representing the coefficients of the $0$-cochain $u$ by sampling the following scalar fields at the nodes of the mesh:
	\begin{align}
		\label{eq:ui}
		&u_{1,i}(x,y) := (i + 1)e^{\sin(x)} + (i+1)^2 e^{\cos(y)}, && i=0,1,2,3,\\
		&u_{2,i}(x,y) := (i + 1)\log(1 + x) + \frac{1}{i+1} \log(1+y), && i=0,1,2,3,\\
		&u_{3,i}(x,y) := x^{i+3} + y^{i+3}, && i=0,1,2,3.
	\end{align} 
	The corresponding source cochains $f$ were computed taking the discrete Laplacian eq.~\eqref{eq:disclap} of these functions. The full dataset is made of the $(u,f)$ pairs. The values of the functions $u_{i,j}$ evaluated at the boundary of the square domain were used to enforce Dirichlet boundary conditions during the minimization of each candidate energy produced in the SR. As initial guesses for the minimization procedures, we took the null $0$-cochain. 
	
	As reported in Table~\ref{tab:hyperparams}, the set of hyperparameter values includes regularization ($\eta=0.1$). With such a set, we obtained a recovery rate of 66\% in the model discovery runs. Notice that the model discovery process is generally effective, even when the exact model is not found, as shown by the decrease of the mean of the fitness of the best individual of each generation, as the evolution proceeds (Fig.~\ref{fig:poisson_plots}A).
	
\begin{figure}
		\centering
		\includegraphics[scale=1]{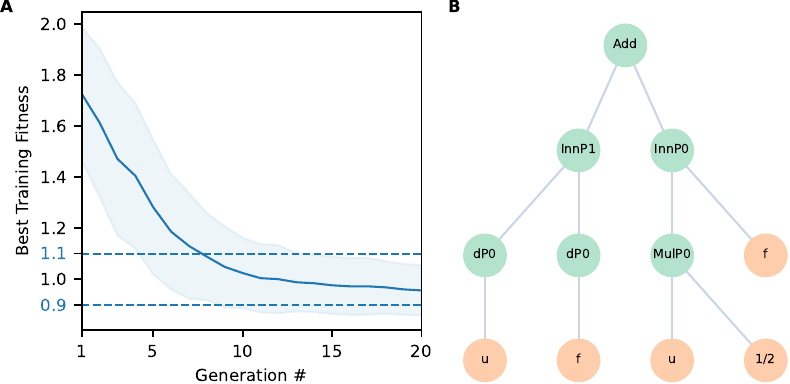}
		
		\caption{(A) Evolution of the mean and standard deviation of the absolute value of the best training fitness over the the final runs $(n=50)$ with $\eta=0.1$ and the initial population consisting of the best 50 individuals found in the runs with $\eta=0$. Dashed horizontal lines mark the fitnesses of individuals having the shortest expressions that represent the energy in eq.~\eqref{eq:discDirichlet}. (B) Tree corresponding to representation of one of the individuals with lowest absolute value of the training fitness (see Table~\ref{tab:primitives} for the meaning of the primitives appearing in the tree).}
		\label{fig:poisson_final}
	\end{figure}

	To assess the effect of regularization on model discovery, we performed 50 additional runs with the best combination of hyperparameters without regularization ($\eta = 0$) found in the grid search (see Table~\ref{tab:grid_search}). In this case, the recovery rate dropped to 60\%, while the length of the tree corresponding to the best individual found at the end of the SR procedure was $39.97\pm 15.53$, which is significantly larger than that (9) of the shortest string corresponding to the energy eq.~\eqref{eq:discDirichlet}. To compensate for such a growth in model complexity, we set $\eta=0.1$ while keeping all the other hyperparameters fixed and ran 50 additional evolutions seeding the initial population with the 50 individuals found at the end of the model discovery runs without regularization. After 20 generations, all runs converged (recovery rate = 100\%) to equivalent expressions of the energy eq.~\eqref{eq:discDirichlet}, with length comprised between 9 and 11 (fitness equal to 0.9 and 1.1, respectively), see Fig.~\ref{fig:poisson_final}A.

	\begin{table}[t!]
		\caption{Discrete energies corresponding to the 4 best individuals obtained in the model discovery runs $(n=50)$ with $\eta=0.1$ and the initial population consisting of the best 50 individuals found in the model discovery with $\eta=0$.}
		\begin{center}
			\begin{tabular}{l l c c c}
				\toprule
				\# & Energy & Training Fit. & Test Fit. & Test MSE\\
				\midrule
				1 & $\langle u, \delta(d(u/2) - f) \rangle$ & 0.9 & 0.9 & $9.8 \cdot 10^{-10}$ \\
				2 & $\langle u, \delta(du) - 2f \rangle$ & 0.9 & 0.9 & $9.8 \cdot 10^{-10}$ \\
				3 & $\langle \star(\star u), \delta(du) - 2f \rangle$ & 1.1 & 1.1 & $9.8 \cdot 10^{-10}$\\
				4 & $\langle du, du  \rangle - \langle f, 2u \rangle$ &  1.1 & 1.1 & $9.8 \cdot 10^{-10}$\\
				\bottomrule
			\end{tabular}
		\end{center}
		\label{tab:bestruns_poisson}
	\end{table}

	\subsection{Euler's Elastica}
	
	\begin{figure}[b!]%
		\centering
		\includegraphics[scale=1]{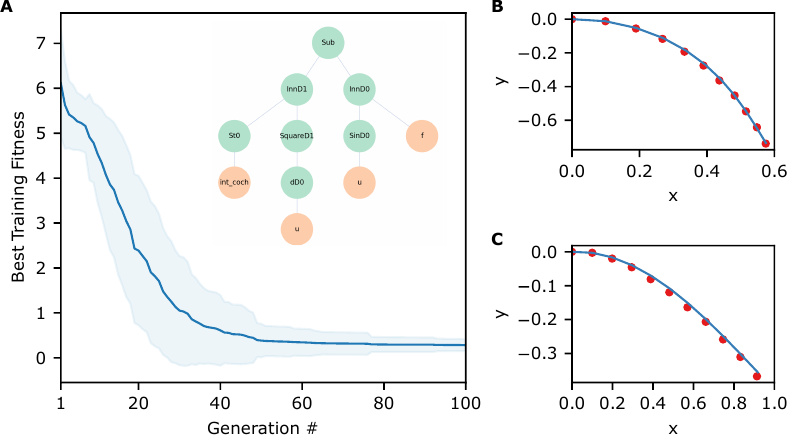}
		\caption{Results for the \textsl{Elastica} problem. (A) Evolution of the mean and standard deviation of the absolute value of the best training fitness over the the model discovery runs $(n=50)$ along with the tree representation of the best individual (\texttt{int\_coch} is $\mathbbm{1}_{\text{int}}$). Comparison between the deformed configurations corresponding to the best individual at the end of the model discovery stage (blue curves) evaluated for \mbox{$P= -45$ N} (B) and $P= -10$ N (C) and the test data (red dots).}
		\label{fig:ela_plots}%
	\end{figure}
	
	Unlike the previous problem, here we emulate a physical experiment involving an elastic cantilever rod subject to an end load by generating synthetic experimental data, instead of starting from an exact solution of a given discrete model. Specifically, we compute numerically the solution of a \textsl{continuous} model for different values of the load, we perturb it with noise and sample it at mesh nodes, to mimic experimental measurements. In the following, $\mathcal{I} = [0,L]\subset \rr$ is a one-dimensional interval that represents the coordinates of the points of the longitudinal axis of the rod in its undeformed, straight configuration, $L$ is the length of the rod, $\tilde{u}(\sigma)$ is the angle formed by the tangent to the axis of the rod and the horizontal direction at the dimensionless arc-length $\sigma := s/L$. The axis of the rod is assumed to be inextensible and its curvature can be readily computed as $\tilde{u}'$, where a prime denotes differentiation with respect to $\sigma$.
	For a constant cross-section rod subject to a vertical end-load at the right end and clamped at the left end, the boundary-value problem for the (dimensionless) continuous \textsl{Euler's Elastica} equation reads \cite{audoly2000elasticity}
	\begin{align}
		\tilde{u}'' + f \cos \tilde{u} = 0  \quad \text{in } [0,1], \quad \tilde{u}(0) = 0, \quad \tilde{u}'(1) = 0, \label{contElastica}
	\end{align}
	and the corresponding variational formulation is
	\begin{equation}
		\min_{\tilde{u}}\ \frac{1}{2}\int_{0}^1 \left(\tilde{u}'\right)^2\,\textrm{d}\sigma - \int_{0}^1 f \sin \tilde{u}  \,\textrm{d}\sigma ,\quad \tilde{u} (0) = 0. \label{contElasticaEnergy}
	\end{equation}
	In eqs.~\eqref{contElastica}-\eqref{contElasticaEnergy}, $f = PL^2/B$ is the dimensionless load parameter, where $P$ is the vertical component of the load applied at the right end and $B$ is the \textsl{bending stiffness} of the rod.  
	
	In the discrete setting, we represent the unknown of the problem as a dual scalar-valued $0$-cochain  $u$, where the value at each dual node is the angle formed by the corresponding primal edge and the horizontal direction. The discrete curvature $k$ may be evaluated by first taking the difference between the angles of two consecutive edges ($d^\star u$), then transferring this information to the primal nodes ($\star d^\star u$) and finally considering only the internal nodes ($\mathbbm{1}_{\text{int}} \odot \star d^\star u$), where $k$ is meaningful. Hence, the discrete equivalent of \eqref{contElasticaEnergy} reads
	\begin{equation}
		\mathcal{E}_{\text{el}}(u) := \frac{1}{2} \langle k, k \rangle - \langle f\mathbbm{1}, \sin u  \rangle, \label{eq:elastica}
	\end{equation}
	where $k = \mathbbm{1}_{\text{int}} \odot \star d^\star u$.
	The 1D mesh for this problem was generated directly by code. Precisely, we generated a $1$-simplicial complex in $[0,1]$ consisting of $11$ uniformly distributed nodes.
	
	To generate the datasets, we considered a rod with length $L=1$ m and bending stiffness $B \approx 7.854\ \textrm{N}\textrm{m}^2$. We computed numerically the solutions of the continuous problem eq.~\eqref{contElastica} for $10$ different values of the force $P \in \{-5,-10,\dots, -45, -50\}$ N and we sampled the $(x,y)$ coordinates of the points of the deformed configurations at the nodes of the $1$-simplicial complex. Finally, we perturbed them with uniformly distributed noise (amplitude equal to $\approx 5\%$ of the order of magnitude of the clean data) and we recovered the arrays of the coefficients of $u$. Linear approximations of these data were used as initial guesses for the minimization of the candidate energies found during the SR. The full dataset consists of the $(u,PL^2)$ pairs.
	
	In a realistic experimental scenario, the load magnitude is prescribed, while the bending stiffness is an unknown material-dependent parameter that must be calibrated to reproduce the measurements. Hence, for each individual of the SR procedure the bending stiffness must be fitted on one solution $\bar{u}$ belonging to the training set. For this purpose, for each candidate energy $\mathcal{E}(u,f)$ we solved the constrained optimization problem
	\begin{align}
		\min_{f \geq 0} \quad & ||u_f - \bar{u}||^2 \qquad
		\textrm{s.t.}\ u_f \in \argmin_{u \text{ : } u(0) = 0} \mathcal{E}(u,f) \label{bilevel}
	\end{align} 
	Then, from the relation $B = PL^2/f$ we recovered the value of $B$ and used it to evaluate the fitness of the individual over the whole training, validation and test sets, as described in Section~\ref{sec:symbreg}. In particular, taking the ground-truth model, $\mathcal{E} \equiv \mathcal{E}_{\text{el}}$, and solving the optimization problem we can identify a value $B_{\text{noise}} \approx 7.4062\ \textrm{N}\textrm{m}^2$ for the bending stiffness that accounts for the noise.
	
	\begin{table}[t!]
		\caption{Discrete energies corresponding to the 4 best individuals obtained in the model discovery runs $(n=50)$ of the \textsl{Elastica} problem using the final values of the hyperparameters reported in Table~\ref{tab:hyperparams}.}
		\begin{center}
				\begin{tabular}{l l c c c c}
					\toprule
					\# & Energy & Training Fit. & Test Fit. & Test MSE & $B$\\
					\midrule
					1 & $\langle \star \mathbbm{1}_{\text{int}}, (d^\star u)^2\rangle - \langle \sin u, f\mathbbm{1}\rangle$ & 0.196 & 0.1939 & 0.0084 & 37.0312 Nm$^2$\\
					2 & $\langle \arccos(-1) \sin u + \delta^\star(\sin(\sin(d^\star u))), u - f\mathbbm{1} \rangle$ & 0.2123 & 0.2247 & 0.0075 & 7.1779 Nm$^2$\\
					3 & $\langle u - f\mathbbm{1}, \delta^\star (\sin(\sin(d^\star u))) + 1/\exp(-1)\sin u \rangle$ & 0.214 & 0.2168 & 0.0067& 6.8262 Nm$^2$\\
					4 & $\langle \star \mathbbm{1}_{\text{int}}, (d^\star u)^2 \rangle - \langle f\mathbbm{1}, \sin u \rangle + 1/2$ & 0.216 & 0.2139 & 0.0084& 37.0312 Nm$^2$\\
					\bottomrule
			\end{tabular}
		\end{center}
		\label{tab:bestruns}
	\end{table}

	Both the average and the standard deviation of the fitness of the best individual of each generation over 50 model discovery runs decrease during the SR evolution (Fig.~\ref{fig:ela_plots}A) and attain the values 0.28 and 0.13, respectively, after 100 generations. Also, the selected models fit the data of the test set well (Fig.~\ref{fig:ela_plots}B-C). These observations confirm the effectiveness of the model discovery process along with the accuracy and generalization capabilities of the recovered models. 
	
	We notice that the first and the last models among top $4$ of Table~\ref{tab:bestruns} are such that $B = B_{\text{noise}}/2h$,	where $h$ is the (primal) edge length. This relation exploits the uniform edge length of the simplicial complex representing the axis of the rod. Therefore, these models coincide with that of eq.~\eqref{eq:elastica} up to addition and/or multiplication with a constant.

	\subsection{Linear Elasticity in 2D}\label{sec:linel}
	
	Differently from the previous benchmarks, this case deals with a vector-valued unknown and thus proves the capability of our framework of handling another class of problems. 
	We collect the dimensionless coordinates of the nodes of the 2-simplex representing the reference configuration of an elastic body in a primal vector-valued $0$-cochain. We assume plane strain conditions. Let $\bm{e}_i, i=1,2$ be the covariant basis for $\mathcal{V}$ made of two edge vectors of the reference configuration of a given $2$-simplex and let $\bm{e}'_i$ be the covariant basis $\mathcal{V}$ made of two edge vectors of the corresponding current configuration of the $2$-simplex. The contravariant bases are indicated with superscript indices. We represent the two-dimensional deformation gradient $\bm{F}$ as a dual (tensor-valued) 0-cochain whose coefficient on a given $2$-simplex is
	\begin{equation}
		\bm{F}(\star \sigma^2) := {\bm{e}'}_i \otimes {\bm{e}}^i. \label{eq:defgrad}
	\end{equation}
	Then, we define the two-dimensional, infinitesimal strain measure as a dual $0$-cochain
	\begin{equation}
		\bm{E} := \frac{1}{2}(\bm{F} + \bm{F}^T) - \bm{I}, \label{eq:strain}
	\end{equation}
	and by applying the constitutive equation for linearly elastic isotropic bodies we obtain the corresponding dimensionless (two-dimensional) stress as a dual $0$-cochain
	\begin{equation}
		\label{eq:stress}
		\bm{S} := 2\bm{E} + \frac{\lambda\_}{\mu\_}\,\text{tr}(\bm{E})\bm{I},
	\end{equation}
	where $\bm{I}$ is the identity dual $0$-cochain and $(\mu\_, \lambda\_)$ are the Lamé moduli (here, we take $\lambda\_ / \mu\_ = 10$). Notice that the second term in the constitutive equation involves the coefficient-wise multiplication between a scalar-valued and a tensor-valued cochain.
	
	In the case of zero body forces and assuming Dirichlet boundary conditions (prescribed positions of the nodes), the potential energy function reads
	\begin{equation}
		\mathcal{E}_{LE} := \frac{1}{2} \langle \bm{S},\bm{E} \rangle. \label{eq:lin_ela_energy}
	\end{equation}
	The goal of this benchmark is recovering eq.~\eqref{eq:lin_ela_energy} as a function of $\bm{F}$. The deformed configuration is given in terms of the current positions of the nodes that minimized eq.~\eqref{eq:lin_ela_energy}. In principle, an admissible candidate energy $\mathcal{E}(\bm{F})$ must be \textsl{frame-indifferent}, \textsl{i.e.}~$\mathcal{E}(\bm{F} + \bm{W}) = \mathcal{E}(\bm{F})$ for any skew-symmetric dual $0$-cochain $\bm{W}$. To enforce such a constraint, we added the penalty term $- (\mathcal{E}(\bm{F} + \tilde{\bm{W}}) - \mathcal{E}(\bm{F}))^2$ to the fitness eq.~\eqref{eq:fitness}, where
	\begin{equation}
		\tilde{\bm{W}} := \sum_{i=0}^{n_2} \begin{bmatrix}
			0 & e\\
			-e & 0
		\end{bmatrix} \star\!{\sigma^{2,i}}.
	\end{equation}
	This addition is clearly necessary but not sufficient. In practice, we observed that all the energies obtained at the end of the model discovery runs are frame-indifferent.
	
	We considered a 2-simplicial complex (dimensionless reference configuration) generated as a Delaunay triangulation of the unit square with $142$ nodes. The full dataset (training, validation and test sets) consists of 20 samples, each including the positions of the nodes in the deformed configurations associated to the problems of uniaxial tension and pure shear. For these problems, the (uniform) strain and stress cochains are:
	\begin{itemize}
		\item \textsl{Uniaxial tension}:
		\begin{equation}
			\bm{S}_{pt} := \sum_{i=0}^{n_2} \begin{bmatrix}
				s & 0\\
				0 & 0
			\end{bmatrix} \star\!\sigma^{2,i}, \quad \bm{E}_{pt} := \sum_{i=0}^{n_2} \begin{bmatrix}
				\epsilon & 0\\
				0 & \ell 
			\end{bmatrix} \star\!\sigma^{2,i},
		\end{equation}
		where $s = 2\epsilon + (\lambda\_/\mu\_)(\epsilon + \ell)$, $\ell = - (\lambda\_/\mu\_)\,\epsilon/(2 + (\lambda\_/\mu\_))$, and $\epsilon \in \{0.01, 0.02, \dots, 0.1\}$;
		\item \textsl{Pure shear}:
		\begin{equation}
			\bm{S}_{ps} := \sum_{i=0}^{n_2} \begin{bmatrix}
				0 & \gamma\\
				\gamma & 0
			\end{bmatrix} \star\!\sigma^{2,i}, \quad \bm{E}_{ps} :=  \sum_{i=0}^{n_2} \frac{1}{2}\begin{bmatrix}
				0 & \gamma\\
				\gamma & 0
			\end{bmatrix} \star\!\sigma^{2,i},
		\end{equation}
		where $\gamma \in \{0.01, 0.02, \dots, 0.1\}$.
	\end{itemize}
	In both cases, the current positions of the nodes can be easily computed from the strain cochain and from eqs.~\eqref{eq:defgrad}-\eqref{eq:strain}. As initial guess for the energy minimization procedures associated to the evaluation of the fitnesses of the candidate energies, we took the reference positions of the nodes.

	\begin{table}[t!]
		\caption{Discrete energies corresponding to the 4 best individuals obtained from the model discovery runs $(n=50)$ of the \textsl{Linear Elasticity} problem using the final values of the hyperparameters reported in Table~\ref{tab:hyperparams}.}
		\begin{center}
				\begin{tabular}{l l c c}
					\toprule
					\# & Energy & Training MSE & Test MSE\\
					\midrule
					1& $\langle \bm{I} - \bm{F}^T, \bm{I} \rangle^2 - 0.1(\langle \bm{I} - \bm{F}^T, \text{sym}(\bm{F}) - \bm{I} \rangle + \langle \bm{I} - \text{sym}(\bm{F}), \text{sym}(\bm{F}) - \bm{I} \rangle)$ & 0 & $1.1672 \cdot 10^{-16}$\\
					2& $\langle -0.5\bm{I} + 0.5 \,\text{sym}(\bm{F}), \text{sym}(\bm{F}) - \bm{I} + (\text{tr}(\bm{F}) - \text{tr}(\bm{I}))(2\text{tr}(\bm{I})\bm{I} + \bm{I})  \rangle$ &  0 & $2.0785 \cdot 10^{-16}$\\
					3& $\langle \bm{I} - \bm{F}^T, \bm{I} - \text{sym}(\bm{F}) + 5\langle \bm{I} - \bm{F}, \bm{I} \rangle \bm{I} \rangle$ & 0 & $3.7309 \cdot 10^{-16}$\\
					4& $\langle \text{sym}(\bm{F}) - \bm{I}, \text{sym}(\bm{F}) - \bm{I} \rangle + 5 \langle \bm{I} - \bm{F}, \bm{I} \rangle^2$ & 0 & $8.7033 \cdot 10^{-16}$\\
					\bottomrule
			\end{tabular}
		\end{center}
		\label{tab:finalruns_elasticity}
	\end{table}
	
	\begin{figure}[t!]%
		\centering
		\includegraphics[scale=1]{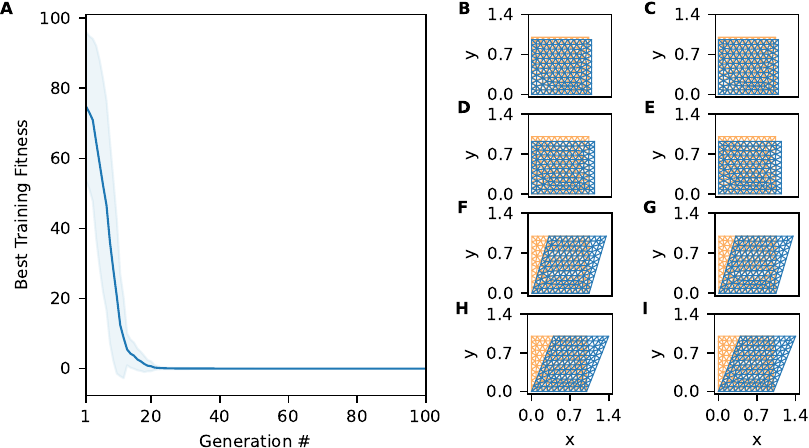}
		\caption{Result for the \textsl{Linear Elasticity} problem. (A) Evolution of the mean and standard deviation of the absolute value of the best training fitness over the model discovery runs $(n=50)$. Comparison between the deformed configurations (B, D) for $\epsilon \in \{0.01, 0.02\}$ (uniaxial tension) and (F, H) for $\gamma \in \{0.06, 0.08\}$ (pure shear) corresponding to the best model at the end of the model discovery stage and the associated test data (C, E) and (G, I), respectively. The magnitude of the displacements has been scaled by a factor of $5$ for clarity. Orange meshes represent the reference configurations.}
		\label{fig:linear_elasticity_plots}%
	\end{figure}
	
	In the model discovery runs performed using the set of hyperparameters of Table~\ref{tab:hyperparams}, we obtained a recovery rate of $92\%$. However, the SR process is always very effective, even in the runs where the correct solution is not attained (Fig.~\ref{fig:linear_elasticity_plots}A). Notice that the second best individual of Table~\ref{tab:finalruns_elasticity} equals $1/2 \, \mathcal{E}_{LE}$. For the remaining energies reported in Table~\ref{tab:finalruns_elasticity}, we can still prove their equivalence to $\mathcal{E}_{LE}$, under specific hypothesis. In the case of homogeneous deformations ($\bm{F}(\star \sigma_{2,i}) = \tilde{\bm{F}}$ for any $i$)
	\begin{align}
		&\langle \bm{I}, \bm{F}^T \rangle^2 = \big(\sum_{i=0}^{n_2} |\sigma_{2,i}| \text{tr}(\bm{F}(\star \sigma_{2,i}))\big)^2 = \text{tr}(\tilde{\bm{F}})^2A^2, \label{eq:proof1}\\ 
		&\langle\text{tr}(\bm{F})\bm{I},  \text{sym}(\bm{F})\rangle = \sum_{i=0}^{n_2} |\sigma_{2,i}| \text{tr}(\bm{F}(\star \sigma_{2,i}))^2 =  \text{tr}(\tilde{\bm{F}})^2A, \label{eq:proof2}\\
		&\langle \bm{I}, \bm{I} \rangle = \sum_{i=0}^{n_2} |\sigma_{2,i}| \text{tr}(\bm{I}) = 2A, \label{eq:proof3}
	\end{align}
	where $A$ is the total area of the $2$-complex.
	Moreover, the following identities hold in general:
	\begin{align}
		&\langle \bm{I} - \bm{F}, \text{sym}(\bm{F})\rangle = \langle \bm{I} - \bm{F}^T, \text{sym}(\bm{F})\rangle = \langle \bm{I} - \text{sym}(\bm{F}), \text{sym}(\bm{F})\rangle, \label{eq:proof4}\\
		&\langle \bm{I}, \bm{F} \rangle = \langle \bm{I}, \bm{F}^T \rangle = \langle \bm{I}, \text{sym}(\bm{F}) \rangle, \label{eq:proof5}\\
		&\langle \text{tr}(\bm{F})\bm{I}, \bm{I} \rangle = 2\langle \bm{I}, \bm{F} \rangle. \label{eq:proof6}
	\end{align}
	Additionally, when $A=1$ as in our benchmark, eqs.~\eqref{eq:proof1}-\eqref{eq:proof3} provide the identities $\langle \bm{I}, \bm{F}^T \rangle^2 = \langle\text{tr}(\bm{F})\bm{I},  \text{sym}(\bm{F})\rangle$ and $\langle \bm{I}, \bm{I} \rangle = 2$.
	These facts, together with eqs.~\eqref{eq:proof4}-\eqref{eq:proof6}, allow to prove the aforementioned equivalences.
	Of course, some of these equivalences do not hold in general (\textit{i.e.}~when deformations are not homogeneous), so the recovered models correspond to different expressions of the potential energy, while having the same accuracy (in terms of MSE) with respect to the datasets. To resolve this ambiguity, we compute the associated energy densities by  dividing the energies by $A$, since the deformations are homogeneous. We find that energies \#1, \#3 and \#4 of Table~\ref{tab:finalruns_elasticity} are not admissible, because their energy densities linearly depend on the area $A$ due to the term in eq.~\eqref{eq:proof1}. However, these expressions may be fixed by interpreting such a term as that in eq.~\eqref{eq:proof2}, so that they all correspond to the exact energy density that was used to generate the dataset. We remark that this \textsl{a posteriori} discussion can be done since our method produces interpretable models that are amenable to analysis. 
	
	We notice that traditional SR can be used when the deformation gradient and the stress are homogeneous. However, its performance is much worse than that of our method, because the former lacks tensor-based operations, which are needed to achieve a compact representation of the energy density. The results of the comparison between the two methods are reported in Appendix~\ref{sec:LEcomparison}, along with an example of a genuine field problem in Linear Elasticity, which clearly cannot be solved \textsl{via} a traditional SR approach. We believe that these simple, yet effective examples demonstrate the advantage of the proposed approach in situations where  conventional SR methods struggle to learn an accurate model.

	\section{Conclusions}
	In this work, we have introduced a new method to learn discrete energies of physical system from experimental data that exploits DEC as an expressive language for physical theories and the interpretability stemming from the SR approach. We have demonstrated its effectiveness by studying three classical problems in Continuum Physics. Our approach suggests that the synergy between geometry and SR leads to physically-consistent models that generalize well. Further, it paves the way to the application of SR to the modeling of a broad class of physical systems.
	
	\paragraph{Limitations and future work} We have developed our framework based on a special class of problems (variational and steady-state). This choice was motivated by the simpler implementation of the numerical routines compared to time-dependent, non-variational problems. To overcome this limitation, we will implement time-integration schemes and, regarding non-variational problems, we will use trees to represent the residual of the system of governing equations, instead of the energy functional.
	
	\section*{Acknowledgments}
	Funded by the European Union (ERC, ALPS, 101039481). Views and opinions expressed are however those of the author(s) only and do not necessarily reflect those of the European Union or the European Research Council Executive Agency. Neither the European Union nor the granting authority can be held responsible for them.
		
	\bibliographystyle{ieeetr}

\begin{thebibliography}{10}
		
		\bibitem{schmidt2009distilling}
		M.~Schmidt and H.~Lipson, ``Distilling free-form natural laws from experimental
		data,'' {\em Science}, vol.~324, no.~5923, pp.~81--85, 2009.
		
		\bibitem{bongard2007automated}
		J.~Bongard and H.~Lipson, ``Automated reverse engineering of nonlinear
		dynamical systems,'' {\em Proceedings of the National Academy of Sciences},
		vol.~104, no.~24, pp.~9943--9948, 2007.
		
		\bibitem{brunton2016discovering}
		S.~L. Brunton, J.~L. Proctor, and J.~N. Kutz, ``Discovering governing equations
		from data by sparse identification of nonlinear dynamical systems,'' {\em
			Proceedings of the National Academy of Sciences}, vol.~113, no.~15,
		pp.~3932--3937, 2016.
		
		\bibitem{udrescu2020ai}
		S.-M. Udrescu and M.~Tegmark, ``{AI Feynman}: A physics-inspired method for
		symbolic regression,'' {\em Science Advances}, vol.~6, no.~16, p.~eaay2631,
		2020.
		
		\bibitem{cranmer2020discovering}
		M.~Cranmer, A.~Sanchez~Gonzalez, P.~Battaglia, R.~Xu, K.~Cranmer, D.~Spergel,
		and S.~Ho, ``Discovering symbolic models from deep learning with inductive
		biases,'' {\em Advances in Neural Information Processing Systems}, vol.~33,
		pp.~17429--17442, 2020.
		
		\bibitem{petersen2019deep}
		B.~K. Petersen, M.~Landajuela, T.~N. Mundhenk, C.~P. Santiago, S.~K. Kim, and
		J.~T. Kim, ``Deep symbolic regression: Recovering mathematical expressions
		from data via risk-seeking policy gradients,'' {\em arXiv preprint
			arXiv:1912.04871}, 2019.
		
		\bibitem{biggio2021neural}
		L.~Biggio, T.~Bendinelli, A.~Neitz, A.~Lucchi, and G.~Parascandolo, ``Neural
		symbolic regression that scales,'' in {\em International Conference on
			Machine Learning}, pp.~936--945, PMLR, 2021.
		
		\bibitem{rudy2017data}
		S.~H. Rudy, S.~L. Brunton, J.~L. Proctor, and J.~N. Kutz, ``Data-driven
		discovery of partial differential equations,'' {\em Science Advances},
		vol.~3, no.~4, 2017.
		
		\bibitem{kaptanoglu2021pysindy}
		A.~A. Kaptanoglu, B.~M. de~Silva, U.~Fasel, K.~Kaheman, A.~J. Goldschmidt,
		J.~L. Callaham, C.~B. Delahunt, Z.~G. Nicolaou, K.~Champion, J.-C. Loiseau,
		{\em et~al.}, ``PySINDY: A comprehensive Python package for robust sparse
		system identification,'' {\em arXiv preprint arXiv:2111.08481}, 2021.
		
		\bibitem{grinspun2006discrete}
		E.~Grinspun, M.~Desbrun, K.~Polthier, P.~Schr{\"o}der, and A.~Stern, ``Discrete
		differential geometry: an applied introduction,'' {\em ACM SIGGRAPH Course},
		vol.~7, no.~1, 2006.
		
		\bibitem{hirani2003discrete}
		A.~N. Hirani, {\em Discrete exterior calculus}.
		\newblock California Institute of Technology, 2003.
		
		\bibitem{tonti2013mathematical}
		E.~Tonti, {\em The mathematical structure of classical and relativistic
			physics}, vol.~10.
		\newblock Springer, 2013.
		
		\bibitem{tonti2014starting}
		E.~Tonti, ``Why starting from differential equations for computational
		physics?,'' {\em Journal of Computational Physics}, vol.~257, pp.~1260--1290,
		2014.
		
		\bibitem{milicchio2008codimension}
		F.~Milicchio, A.~DiCarlo, A.~Paoluzzi, and V.~Shapiro, ``A codimension-zero
		approach to discretizing and solving field problems,'' {\em Advanced
			Engineering Informatics}, vol.~22, no.~2, pp.~172--185, 2008.
		
		\bibitem{dicarlo2009chain}
		A.~DiCarlo, F.~Milicchio, A.~Paoluzzi, and V.~Shapiro, ``Chain-based
		representations for solid and physical modeling,'' {\em IEEE Transactions on
			Automation Science and Engineering}, vol.~6, no.~3, pp.~454--467, 2009.
		
		\bibitem{dicarlo2009discrete}
		A.~DiCarlo, F.~Milicchio, A.~Paoluzzi, and V.~Shapiro, ``Discrete physics using
		metrized chains,'' in {\em 2009 SIAM/ACM joint conference on geometric and
			physical modeling}, pp.~135--145, 2009.
		
		\bibitem{behandish2022ai}
		M.~Behandish, J.~Maxwell~III, and J.~de~Kleer, ``{AI Research Associate for
			Early-Stage Scientific Discovery},'' {\em arXiv preprint arXiv:2202.03199},
		2022.
		
		\bibitem{sahoo2018learning}
		S.~Sahoo, C.~Lampert, and G.~Martius, ``Learning equations for extrapolation
		and control,'' in {\em International Conference on Machine Learning},
		pp.~4442--4450, PMLR, 2018.
		
		\bibitem{cranmer2023interpretable}
		M.~Cranmer, ``Interpretable machine learning for science with PySR and
		SymbolicRegression.jl,'' {\em arXiv preprint arXiv:2305.01582}, 2023.
		
		\bibitem{grady2010discrete}
		L.~J. Grady and J.~R. Polimeni, {\em Discrete calculus: Applied analysis on
			graphs for computational science}, vol.~3.
		\newblock Springer, 2010.
		
		\bibitem{koza1994genetic}
		J.~R. Koza, {\em Genetic Programming II: Automatic Discovery of Reusable
			Programs}.
		\newblock Brandford Books, 1994.
		
		\bibitem{o2009riccardo}
		M.~O'Neill, R.~Poli, W.~B. Langdon, and N.~F. McPhee, {\em A Field Guide to
			Genetic Programming}.
		\newblock Springer, 2009.
		
		\bibitem{de2016evolutionary}
		K.~De~Jong, {\em Evolutionary computation: a unified approach}.
		\newblock MIT Press, 2006.
		
		\bibitem{jax2018github}
		J.~Bradbury, R.~Frostig, P.~Hawkins, M.~J. Johnson, C.~Leary, D.~Maclaurin,
		G.~Necula, A.~Paszke, J.~Vander{P}las, S.~Wanderman-{M}ilne, and Q.~Zhang,
		``{JAX}: composable transformations of {P}ython+{N}um{P}y programs,'' 2018.
		
		\bibitem{Biscani2020}
		F.~Biscani and D.~Izzo, ``A parallel global multiobjective framework for
		optimization: pagmo,'' {\em Journal of Open Source Software}, vol.~5, no.~53,
		p.~2338, 2020.
		
		\bibitem{DEAP_JMLR2012}
		F.-A. Fortin, F.-M. {De Rainville}, M.-A. Gardner, M.~Parizeau, and C.~Gagn\'e,
		``{DEAP}: Evolutionary algorithms made easy,'' {\em Journal of Machine
			Learning Research}, vol.~13, pp.~2171--2175, 2012.
		
		\bibitem{ray}
		P.~Moritz, R.~Nishihara, S.~Wang, A.~Tumanov, R.~Liaw, E.~Liang, M.~Elibol,
		Z.~Yang, W.~Paul, M.~I. Jordan, and I.~Stoica, ``Ray: A {Distributed
			Framework for Emerging AI Applications},'' OSDI'18, (USA), pp.~561--577,
		USENIX Association, 2018.
		
		\bibitem{podio2019continuum}
		P.~Podio-Guidugli, {\em Continuum Thermodynamics}.
		\newblock Springer, 2019.
		
		\bibitem{audoly2000elasticity}
		B.~Audoly and Y.~Pomeau, {\em Elasticity and geometry}.
		\newblock Oxford University Press, 2018.
		
		\bibitem{gurtin1982introduction}
		M.~E. Gurtin, {\em An Introduction to Continuum Mechanics}, vol.~158.
		\newblock AcademicPress, 1981.
		
		\bibitem{evans2010partial}
		L.~Evans, {\em Partial Differential Equations}.
		\newblock Graduate studies in mathematics, American Mathematical Society, 2010.
		
		\bibitem{desbrun2005discrete}
		M.~Desbrun, A.~N. Hirani, M.~Leok, and J.~E. Marsden, ``Discrete exterior
		calculus,'' {\em arXiv preprint arXiv:math/0508341}, 2005.
		
		\bibitem{geuzaine2009gmsh}
		C.~Geuzaine and J.-F. Remacle, ``Gmsh: {A 3-D} finite element mesh generator
		with built-in pre-and post-processing facilities,'' {\em International
			journal for numerical methods in engineering}, vol.~79, no.~11,
		pp.~1309--1331, 2009.
		
		\bibitem{schulz2020convergence}
		E.~Schulz and G.~Tsogtgerel, ``Convergence of discrete exterior calculus
		approximations for Poisson problems,'' {\em Discrete \& Computational
			Geometry}, vol.~63, pp.~346--376, 2020.
		
		\bibitem{frankel2011geometry}
		T.~Frankel, {\em The Geometry of Physics: an Introduction}.
		\newblock Cambridge University Press, 2011.
		
	\end{thebibliography}

	\newpage
	
	\appendix
	\section*{Appendix}
	\numberwithin{equation}{subsection}
	\addcontentsline{toc}{section}{Appendices}
	\renewcommand{\thesubsection}{\Alph{subsection}}
	\counterwithin{figure}{subsection}
	\counterwithin{table}{subsection}

	\subsection{Mathematical details on DEC}
	\label{DEC details}
	
	\paragraph{Simplicial Complex.} A \textsl{$p$-simplex} is the convex hull of $p+1$ independent points in $\mathcal{E}^N$. A collection of simplices defines a \textsl{simplicial complex} $K$ if it satisfies the following conditions:
	\begin{itemize}
		\setlength\itemsep{0.01em}
		\item every lower-order simplex of a given simplex in $K$ is in $K$;
		\item the intersection of any two simplices is either empty or a boundary element of both simplices.
	\end{itemize}

	\paragraph{Chains and cochains.} A (scalar-valued) \textsl{$p$-chain} is a collection of values, one for each of the $p$-simplices of $K$. Each $p$-simplex corresponds to a \textsl{basic chain} $\sigma_{p,i}$ (\textsl{i.e.} a chain assigning $1$ to a given simplex and $0$ to the others) and we can use the same notation to indicate both the simplex and the corresponding basic chain. The $p$-chains group is denoted by $\mathcal{C}_p(K, \mathbb{R})$. A $p$-chain $c_p$ can be written in an unique way as a finite sum of basic chains:
	\begin{align}
		c_p = \sum_{i=1}^{n_p} c_p(\sigma_{p,i}) \sigma_{p,i},
	\end{align}
	where $n_p$ is the number of $p$-simplices of $K$, $c_p(\sigma_{p,i}) \in \mathbb{R}$, and $n_p$ is the number of $p$-simplices. In computations, a chain can be represented by a vector where the $i$-th entry is the value of the chain on the $i$-th simplex.
	
	A $p$-\textsl{cochain} is a linear functional that maps chains to scalars. Analogously to chains, we can represent a cochain $c^p$ in terms of \textsl{basic cochains}
	\begin{align}
		c^p = \sum_{i=1}^{n_p} c^p(\sigma^{p,i}) \sigma^{p,i},
	\end{align}
	where $c^p(\sigma^{p,i})$ are coefficients to be chosen in an appropriate space and $\sigma^{p,i}$ is the $i$-th basic $p$-cochain, \textsl{i.e.}~the cochain that assigns $1$ to the $i$-th basic chain, which we can again identify with the $i$-th simplex, and 0 to the others. The space of scalar-valued $p$-cochains on a simplicial complex $K$ is denoted by $\mathcal{C}^p(K, \mathbb{R})$, while the spaces of vector-valued and tensor-valued cochains are denoted by $\mathcal{C}^p(K, \mathcal{V})$ and $\mathcal{C}^p(K, \text{Lin})$, respectively. We indicate any of these spaces with $\mathcal{C}^p(K, \cdot)$.
	We will use bold symbols for vector-valued and tensor-valued cochains and for their coefficients $\bm{c}^p(\sigma^p_i) \in \mathcal{V}$ and $\bm{C}^p(\sigma^p_i) \in \text{Lin}$, respectively. In computations, a cochain can be represented by a multi-dimensionl array, where each row contains its value(s) associated to a given simplex. 
	
	A $p$-cochain can be readily evaluated on a $p$-chain by exploiting the duality between basic chains and basic cochains. We denote such an evaluation by the pairing
	\begin{align}
		\llangle c^p, c_p \rrangle := \sum_{i=1}^{n_p} c^p(\sigma^{p,i}) c_p(\sigma_{p,i}). \label{eq:pairing}
	\end{align}
	with $\llangle \sigma^{p,i}, \sigma_{p,j} \rrangle = \delta^i_j$. Notice that $\llangle c^p, \sigma_{p,i} \rrangle = c^p(\sigma^{p,i})$ and $\llangle \sigma^{p,i}, c_p \rrangle = c_p(\sigma_{p,i})$.
	This operation is a discrete preliminary to \textsl{integration} when regarding cochains as densities with respect to the measures imparted to cells by real-valued chains \cite{dicarlo2009discrete}.
	
	\paragraph{Boundary and coboundary.} The \textsl{boundary} of a $p$-simplex is the union of its $(p-1)$-faces, that is, $(p-1)$-simplices consisting of subsets of the vertices of the $p$-simplex. Formally, the \textsl{boundary operator} is a linear map $\partial: \mathcal{C}_p(K, \rr) \rightarrow \mathcal{C}_{p-1}(K, \rr)$ defined by
	\begin{align}
		\partial \sigma_p = \sum_{i = 0}^p (-1)^i [\sigma_{0,0}, \dots, \hat{\sigma}_{0,i}, \dots, \sigma_{0,p}],
	\end{align}
	where $\sigma_p = [\sigma_{0,0}, \dots, \sigma_{0,p}]$ is a generic $p$-simplex and $[\sigma_{0,0}, \dots, \hat{\sigma}_{0,i}, \dots, \sigma_{0,p}]$ is the $(p-1)$-simplex obtained omitting the $i$-th vertex. The boundary operator encodes fundamental topological information about the simplicial complex and allows to define the \textsl{coboundary operator} $d: \mathcal{C}^{p-1}(K, \rr) \rightarrow \mathcal{C}^p(K, \rr)$ as its adjoint with respect to the pairing defined in eq.~\eqref{eq:pairing}:
	\begin{align}
		\llangle d c^{p-1}, c_p \rrangle = \llangle c^{p-1}, \partial c_p \rrangle.
	\end{align}
	The boundary operator $\partial$ may be represented by the \textsl{incidence matrix} \cite{tonti2013mathematical, grady2010discrete}, while the coboundary operator corresponds to its transpose. Notice that there is one (co)boundary matrix per simplex dimension.
	
	\paragraph{Dual complex.} Given a simplicial complex $K$ of dimension $n$, we can define its \textsl{dual} $\star K$ by associating each $p$-simplex $\sigma_p \in K$ with its dual $(n-p)$-simplex $\star \sigma_{(n-p)}$, following a specific rule (see Remark 2.5.1 in \cite{hirani2003discrete}).
	In this work, we use the \textsl{circumcentric (or Voronoi) dual}. Once the orientation of the primal is chosen, that of the dual is induced, and the dual boundary and coboundary $d^\star$ account for such orientation (see \cite{grinspun2006discrete, hirani2003discrete, schulz2020convergence} for more details).
	
	\paragraph{Discrete Hodge star.}  The \textsl{diagonal discrete Hodge star}  $\star: \mathcal{C}^p(K, \cdot) \rightarrow \mathcal{C}^{n-p}(\star K, \cdot)$ is defined as follows \cite{hirani2003discrete, grinspun2006discrete}:
	\begin{align}
		\frac{1}{|\star\!\sigma_p|}\star\!c^p(\star \sigma^p)  := \frac{1}{|\sigma_p|} c^p(\sigma^p).
	\end{align}
	Here, $\star \sigma^p$ is the basic dual $(n-p)$-cochain corresponding to $\sigma^p$.
	
	\paragraph{Inner product, codifferential and discrete Laplacian.} We define the $L^2$ inner product on cochains as
	\begin{align}
		\langle c_1^p, c_2^{p} \rangle := \sum_{i=1}^{n_p} c_1^p(\sigma^{p,i}) \cdot \star c_2^{p}(\star \sigma^{p,i}),
	\end{align}
	with $c_1^p, c^p_2 \in \mathcal{C}^p(K, \cdot)$, where the dot denotes the standard multiplication between scalars or the standard Euclidean inner product between vectors or tensors. 
	
	Further, with respect to this inner product, we may introduce the adjoint of the coboundary operator, the \textsl{discrete codifferential operator} $\delta$, which maps \textsl{primal} scalar-valued $p$-cochains into \textsl{primal} scalar-valued $(p-1)$-cochains, as
	\begin{align}
		\delta := (-1)^{n(p-1) + 1} \star d^\star \star,
	\end{align}
	by requiring that $\langle d\alpha, \beta  \rangle = \langle \alpha, \delta \beta \rangle$, for any $\alpha \in \mathcal{C}^{k-1}(K, \rr)$, $\beta \in \mathcal{C}^{k}(K, \rr)$.  In the previous statement, we write $\star$ for both the Hodge star and its inverse and $d^\star$ is the dual coboundary operator. The dual inner product and dual codifferential $\delta^\star$ can be analogously defined.
	
	In order to deal with the Poisson equation (see Section~\ref{sec:Poisson}), we introduce the \textsl{discrete Laplace-de Rham operator}:
	\begin{align}
		\label{eq:disclap}
		\Delta := d\delta + \delta d.
	\end{align}
	This is the discrete analogue of the Laplace-de Rham operator on differential forms, which coincides with minus the standard Laplacian of a scalar field \cite{frankel2011geometry}.
	
	\paragraph{Cochain functions.} Finally, we extend functions operating on scalar fields (such as $\sin, \cos, \exp, \log$) to cochains by component-wise application. In particular, the component-wise multiplication between scalar-valued cochains will be denoted by the symbol $\odot$.
	
	\newpage
	
	\subsection{List of primitives for the SR used in the numerical experiments}
	
	\begin{table}[!h]
		\caption{List of primitives used in the experiments of Section~\ref{sec:experiments}. \texttt{Add}: $+$; \texttt{Sub}: $-$; \texttt{MulF}: $\cdot$ ; \texttt{Div}: $\divisionsymbol$. \texttt{SinF}: $\sin$; \texttt{ArcsinF}: $\arcsin$; \texttt{CosF}: $\cos$; \texttt{ArccosF}: $\arccos$; \texttt{ExpF}: $\exp$; \texttt{LogF}: $\log$;  \texttt{InvF}: inversion of a float. \texttt{SqrtF}: square root; \texttt{SquareF}: square. \texttt{d}: coboundary; \texttt{del}: codifferential; \texttt{Sin}: sin of a cochain; \texttt{Arcsin}: arcsin of a cochain; \texttt{Cos}: cos of a cochain; \texttt{Arccos}: arccos of a cochain; \texttt{Exp}: exp of a cochain; \texttt{Log}: log of a cochain. \texttt{St}: Hodge star; \texttt{InvSt}: inverse Hodge star. \texttt{Sqrt}: square root of a cochain; \texttt{Square}: square of a cochain.
			\texttt{tran}: transpose of a tensor-valued cochain; \texttt{sym}: symmetric part of a tensor-valued cochain. \texttt{tr}: trace of a tensor-valued cochain. 
			\texttt{Mul}: multiplication between a scalar and a cochain. \texttt{Mulv}: multiplication between a scalar-valued cochain and a tensor-valued cochain. \texttt{InvMul}: multiplication between the inverse of a scalar and a cochain. \texttt{Inn}: inner product between cochains of the same type (either primal or dual). \texttt{AddC}: addition of cochains; \texttt{SubC}: subtraction of cochains. \texttt{CochMul}: component-wise cochain multiplication. In all the operations on cochains \texttt{X = (P,D)}, where \texttt{P} means `primal' and \texttt{D} means `dual'; \texttt{J = (0,\dots,$n$)}; \texttt{R = (S,T)}, where \texttt{S} means `scalar-valued' and \texttt{T} means `tensor-valued'.}
		\begin{center}
			\resizebox{\columnwidth}{!}{
				\begin{tabular}{l c c c c c}
					\toprule
					Primitive names & Input types & Return type & \multicolumn{3}{c}{Problem}\\
					& & & \textsl{Poisson} & \textsl{Elastica} & \textsl{Linear Elasticity}\\
					\midrule
					\rowcolor{Gainsboro!60}
					\texttt{Add}, \texttt{Sub}, \texttt{MulF}, \texttt{Div} &  (\texttt{float}, \texttt{float}) & \texttt{float}  & \checkmark & \checkmark & \checkmark\\
					\texttt{SinF}, \texttt{ArcsinF}, \texttt{CosF}, \texttt{ArccosF}, \texttt{ExpF}, \texttt{LogF} \texttt{InvF}  & \texttt{float}& \texttt{float} & \checkmark & \checkmark & \xmark  \\
					\rowcolor{Gainsboro!60}
					\texttt{SqrtF}, \texttt{SquareF} & \texttt{float}& \texttt{float} &\checkmark &\checkmark &\checkmark \\
					\texttt{dXJS}, \texttt{delXJS}, \texttt{SinXJS}, \texttt{ArcsinXJS}, &  & & & & \\
					\texttt{CosXJS}, \texttt{ArccosXJS}, \texttt{ExpXJS}, \texttt{LogXJS} & \multirow{-2}{*}{\texttt{CochainXJS}} & \multirow{-2}{*}{\texttt{CochainXJS}} & \multirow{-2}{*}{\checkmark} & \multirow{-2}{*}{\checkmark} & \multirow{-2}{*}{\xmark}\\
					\rowcolor{Gainsboro!60}
					\texttt{StJR}, \texttt{InvStJR} & \texttt{CochainXJR} & \texttt{CochainXJR} & \checkmark & \checkmark & \checkmark\\
					\texttt{SqrtXJR}, \texttt{SquareXJR} & \texttt{CochainXJR} & \texttt{CochainXJR}& \checkmark & \checkmark & \xmark\\
					\rowcolor{Gainsboro!60}
					\texttt{tranXJT}, \texttt{symXJT} & \texttt{CochainXJT} & \texttt{CochainXJT} & \xmark & \xmark & \checkmark\\
					\texttt{trXJT} & \texttt{CochainXJT} & \texttt{CochainXJS} &\xmark &\xmark &\checkmark\\
					\rowcolor{Gainsboro!60}
					\texttt{MulXJR} & (\texttt{CochainXJR}, \texttt{float}) & \texttt{CochainXJR} & \checkmark & \checkmark & \checkmark \\
					\texttt{MulvXJ}& (\texttt{CochainXJS}, \texttt{CochainXJT}) & \texttt{CochainXJT} & \xmark & \xmark & \checkmark\\
					\rowcolor{Gainsboro!60}
					\texttt{InvMulXJS} & (\texttt{CochainXJS}, \texttt{float}) & \texttt{CochainXJS}& \checkmark & \checkmark & \xmark \\
					\texttt{InnXJR} & (\texttt{CochainXJR}, \texttt{CochainXJR})  & \texttt{float} & \checkmark & \checkmark & \checkmark\\
					\rowcolor{Gainsboro!60}
					\texttt{AddCXJR}, \texttt{SubCXJR} & (\texttt{CochainXJR}, \texttt{CochainXJR})  & \texttt{CochainXJR}& \checkmark & \checkmark& \checkmark\\
					\texttt{CochMulXJS} & (\texttt{CochainXJS}, \texttt{CochainXJS})  & \texttt{CochainXJS}& \checkmark & \checkmark & \xmark\\
					\bottomrule
			\end{tabular}}
		\end{center}
		\label{tab:primitives}
	\end{table}
	
	\subsection{Details on the selection of the hyperparameters for the SR}
	\label{sec:gridsearch}

	\begin{table}[t!]
		\caption{Grid search results. Here, MM = \textsl{Mixed Mutation}, ST = \textsl{Stochastic Tournament}. In bold, the best validation scores for the three problems.}
		\begin{center}
			\resizebox{\columnwidth}{!}{
				\begin{tabular}{c c c c c c c c c c}
					\toprule
					Number of individuals & (Crossover, Mutation) prob. & MM prob. & ST prob. & \multicolumn{3}{c}{Regularization} & \multicolumn{3}{c}{Validation MSE (recovery rate)}\\
					& & & & \textsl{Poisson} & \textsl{Elastica} & \textsl{Linear Elasticity} & \textsl{Poisson} & \textsl{Elastica} & \textsl{Linear Elasticity}\\
					\midrule
					1000 & (0, 1) & (0.7,0.2,0.1) & (1, 0) & 0.1 & 0.05 & 0.01 & 5.0714 $\pm$ 4.5954 (20\%) & 0.0784 $\pm$ 0.0999& 1.2424 $\cdot 10^{-5}$ $\pm$ 1.1649 $\cdot 10^{-5}$ (40)\%\\
					1000 & (0, 1) & (0.8,0.2,0) & (1, 0) & 0.1 & 0.05 & 0.01 & 4.2284 $\pm$ 3.8019 (20\%)&  0.0172 $\pm$ 0.0118 & 8.4039 $\cdot 10^{-6}$ $\pm$ 9.7148 $\cdot 10^{-6}$ (20\%)\\
					2000 & (0, 1) & (0.7,0.2,0.1) & (1, 0) & 0.1 & 0.05 & 0.01 & 3.0376 $\pm$ 3.4102 (20\%)& 0.0802 $\pm$ 0.1385 & 1.3233 $\cdot 10^{-5}$ $\pm$ 8.9749 $\cdot 10^{-5}$ (40\%)\\
					2000 & (0, 1) & (0.8,0.2,0) & (1, 0) & 0.1 & 0.05 & 0.01 & 0.4954 $\pm$ 0.4255 (40\%)&  0.0197 $\pm$ 0.0121 & 5.6222 $\cdot 10^{-6}$ $\pm$ 4.9649 $\cdot 10^{-6}$ (0\%)\\
					1000 & (0.2, 0.8) & (0.7,0.2,0.1) & (1, 0) & 0.1 & 0.05 & 0.01 & 1.7346 $\pm$ 3.4561 (60\%)& 0.2475 $\pm$ 0.2159 & 4.3172 $\cdot 10^{-5}$ $\pm$ 7.8832 $\cdot 10^{-5}$ (60\%)\\
					1000 & (0.2, 0.8) & (0.8,0.2,0) & (1, 0) & 0.1 & 0.05 & 0.01 & 0.0250 $\pm$ 0.0501 (80\%)& 0.1190 $\pm$ 0.1741  & 1.6539 $\cdot 10^{-5}$ $\pm$ 1.0686 $\cdot 10^{-5}$ (0\%)\\
					2000 & (0.2, 0.8) & (0.7,0.2,0.1) & (1, 0) & 0.1 & 0.05 & 0.01 & 1.5245 $\pm$ 2.6260 (60\%)& 0.1049 $\pm$ 0.1182 & 5.3876 $\cdot 10^{-6}$ $\pm$ 6.7293 $\cdot 10^{-6}$ (60\%)\\
					2000 & (0.2, 0.8) & (0.8,0.2,0) & (1, 0) & 0.1 & 0.05 & 0.01 & 0.0947 $\pm$ 0.1888 (60\%)& 0.0228 $\pm$ 0.0208 & 5.4618 $\cdot 10^{-6}$ $\pm$ 6.8377 $\cdot 10^{-6}$ (40\%)\\
					1000 & (0, 1) & (0.7,0.2,0.1) & (1, 0) & 0 &0.01 & 0& 2.3072 $\pm$ 2.6184 (20\%)& 0.1752 $\pm$ 0.1992 & \textbf{6.8078 $\bm{\cdot 10^{-12}}$ $\pm$ 1.1160 $\bm{\cdot 10^{-12}}$ (100\%)}\\
					1000 & (0, 1) & (0.8,0.2,0) & (1, 0) & 0 &0.01 & 0& 1.1878 $\pm$ 0.9544 (20\%)& 0.1896 $\pm$ 0.2186 & 1.4765 $\cdot 10^{-5}$ $\pm$ 2.8894 $\cdot 10^{-5}$ (20\%)\\
					2000 & (0, 1) & (0.7,0.2,0.1) & (1, 0) & 0 &0.01 & 0& 0.5699 $\pm$ 0.6354 (20\%)& 0.2047 $\pm$ 0.2529  & \textbf{6.0695 $\bm{\cdot 10^{-15}}$ $\pm$ 1.1909 $\bm{\cdot 10^{-15}}$ (100\%)}\\
					2000 & (0, 1) & (0.8,0.2,0) & (1, 0) & 0 &0.01 & 0& 0.2582 $\pm$ 0.3603 (40\%)& 0.0583 $\pm$ 0.0787 & 4.1890 $\cdot 10^{-5}$ $\pm$ 8.3782 $\cdot 10^{-5}$ (80\%)\\
					1000 & (0.2, 0.8) & (0.7,0.2,0.1) & (1, 0) & 0 &0.01 & 0& 1.6456 $\pm$ 2.9635 (60\%)& 0.0711 $\pm$ 0.1150 & 1.7215 $\cdot 10^{-4}$ $\pm$ 3.1495 $\cdot 10^{-4}$ (40\%)\\
					1000 & (0.2, 0.8) & (0.8,0.2,0) & (1, 0) & 0 &0.01 & 0& 0.4365 $\pm$ 0.4468 (40\%)& 0.0262 $\pm$ 0.0214 & 2.5231 $\cdot 10^{-8}$ $\pm$ 5.0461 $\cdot 10^{-8}$ (80\%)\\
					2000 & (0.2, 0.8) & (0.7,0.2,0.1) & (1, 0) & 0 &0.01 & 0& 1.1467 $\pm$ 1.9329 (60\%)& 0.0795 $\pm$ 0.1396 & 8.0971 $\cdot 10^{-4}$ $\pm$ 1.4177 $\cdot 10^{-4}$ (40\%)\\
					2000 & (0.2, 0.8) & (0.8,0.2,0) & (1, 0) & 0 &0.01 & 0& 0.0474 $\pm$ 0.0947 (80\%)& \textbf{0.0157 $\pm$ 0.0073} & 3.7003 $\cdot 10^{-8}$ $\pm$ 7.3999 $\cdot 10^{-8}$ (80\%)\\
					1000 & (0, 1) & (0.7,0.2,0.1) & (0.7, 0.3) & 0.1 & 0.05 & 0.01 & 1.4862 $\pm$ 2.2846 (60\%)& 0.0496 $\pm$ 0.0681 & 1.6212 $\cdot 10^{-4}$ $\pm$ 3.0510 $\cdot 10^{-4}$ (40\%)\\
					1000 & (0, 1) & (0.8,0.2,0) & (0.7, 0.3) & 0.1 & 0.05 & 0.01 & 2.0693 $\pm$ 2.0785 (20\%)& 0.1094 $\pm$ 0.1353 & 4.1410 $\cdot 10^{-5}$ $\pm$ 7.4200 $\cdot 10^{-5}$ (40\%)\\
					2000 & (0, 1) & (0.7,0.2,0.1) & (0.7, 0.3) &  0.1 & 0.05 & 0.01 & 1.3861 $\pm$ 2.3857 (60\%)& 0.1339 $\pm$ 0.1521 & \textbf{2.5238 $\bm{\cdot 10^{-14}}$ $\pm$ 2.7976 $\bm{\cdot 10^{-14}}$ (100\%)}\\
					2000 & (0, 1) & (0.8,0.2,0) & (0.7, 0.3) &  0.1 & 0.05 & 0.01 & 0.3846 $\pm$ 0.2734 (0\%)& 0.2299 $\pm$ 0.3228 & 9.0010 $\cdot 10^{-6}$ $\pm$ 6.1403 $\cdot 10^{-6}$ (20\%)\\
					1000 & (0.2, 0.8) & (0.7,0.2,0.1) & (0.7, 0.3) & 0.1 & 0.05 & 0.01 & 3.3625 $\pm$ 3.6303 (40\%)& 0.0497 $\pm$ 0.0399 & 1.4959 $\cdot 10^{-5}$ $\pm$ 1.0025 $\cdot 10^{-5}$ (20\%)\\
					1000 & (0.2, 0.8) & (0.8,0.2,0) & (0.7, 0.3) & 0.1 & 0.05 & 0.01 & 1.8037 $\pm$ 3.1416 (40\%)& 0.0804 $\pm$ 0.0770 & 6.8314 $\cdot 10^{-6}$ $\pm$ 5.5778 $\cdot 10^{-6}$ (40\%)\\
					2000 & (0.2, 0.8) & (0.7,0.2,0.1) & (0.7, 0.3) & 0.1 & 0.05 & 0.01 & 1.9301 $\pm$ 2.6795 (60\%)&  0.0663 $\pm$ 0.0615 & 6.1977 $\cdot 10^{-6}$ $\pm$ 7.5906 $\cdot 10^{-6}$ (60\%)\\
					2000 & (0.2, 0.8) & (0.8,0.2,0) & (0.7, 0.3) & 0.1 & 0.05 & 0.01 & \textbf{0.000 $\pm$ 0.000 (100\%)}& 0.0660 $\pm$ 0.0593 & 7.7641 $\cdot 10^{-6}$ $\pm$ 6.4927 $\cdot 10^{-6}$ (40\%)\\
					1000 & (0, 1) & (0.7,0.2,0.1) & (0.7, 0.3) & 0 &0.01 & 0& 2.1301 $\pm$ 2.4017 (20\%)& 0.3302 $\pm$ 0.3987 & 4.5915 $\cdot 10^{-7}$ $\pm$ 9.1827 $\cdot 10^{-7}$ (80\%)\\
					1000 & (0, 1) & (0.8,0.2,0) & (0.7, 0.3) & 0 &0.01 & 0&  2.9042 $\pm$ 2.7186 (20\%)&  0.0588 $\pm$ 0.0917 & 7.7295 $\cdot 10^{-7}$ $\pm$ 1.1600 $\cdot 10^{-6}$ (60\%)\\
					2000 & (0, 1) & (0.7,0.2,0.1) & (0.7, 0.3) & 0 &0.01 & 0& 0.0450 $\pm$ 0.0901 (80\%)& 0.2218 $\pm$ 0.3032 & 5.0241 $\cdot 10^{-6}$ $\pm$ 9.6092 $\cdot 10^{-6}$ (40\%)\\
					2000 & (0, 1) & (0.8,0.2,0) & (0.7, 0.3) & 0 &0.01 & 0& 0.3978 $\pm$ 0.5296 (60\%)&  0.1401 $\pm$ 0.2211 & \textbf{9.0573 $\bm{\cdot 10^{-16}}$ $\pm$ 4.3205 $\bm{\cdot 10^{-16}}$ (100\%)}\\
					1000 & (0.2, 0.8) & (0.7,0.2,0.1) & (0.7, 0.3) & 0 &0.01 & 0&  0.8872 $\pm$ 1.3022 (40\%)& 0.0772 $\pm$ 0.1239 & \textbf{2.6749 $\bm{\cdot 10^{-11}}$ $\pm$ 5.3488 $\bm{\cdot 10^{-11}}$ (100\%)}\\
					1000 & (0.2, 0.8) & (0.8,0.2,0) & (0.7, 0.3) & 0 &0.01 & 0& 2.4462 $\pm$ 2.9052 (40\%)& 0.0276 $\pm$ 0.0267 & 4.3438 $\cdot 10^{-7}$ $\pm$ 8.6875 $\cdot 10^{-7}$ (80\%)\\
					2000 & (0.2, 0.8) & (0.7,0.2,0.1) & (0.7, 0.3) & 0 &0.01 & 0& 0.1696 $\pm$ 0.3391 (80\%)& 0.1049 $\pm$ 0.1089 & \textbf{4.9220 $\bm{\cdot 10^{-15}}$ $\pm$ 8.6064 $\bm{\cdot 10^{-15}}$ (100\%)}\\
					2000 & (0.2, 0.8) & (0.8,0.2,0) & (0.7, 0.3) & 0 &0.01 & 0& 0.6313 $\pm$ 1.2626 (80\%)&  0.0282 $\pm$ 0.0111 & 2.9903 $\cdot 10^{-7}$ $\pm$ 5.9782 $\cdot 10^{-7}$ (80\%)\\
					\bottomrule
			\end{tabular}}
		\end{center}
		\label{tab:grid_search}
	\end{table}

	We report in Table~\ref{tab:grid_search} the results of the grid search carried out for each problem. For \textsl{Poisson} and \textsl{Linear Elasticity}, we used the recovery rate as the performance measure, while for \textsl{Elastica} we took the MSE on the validation set, since the data are noisy. For each combination of hyperparameters we performed $5$ runs. In particular, for \textsl{Linear Elasticity}, we selected the set of hyperparameters among those (5) leading to a $100\%$ recovery rate in the grid search after carrying out $50$ final model discovery runs. We reported in Table~\ref{tab:hyperparams} the set of hyperparameters corresponding to the best recovery rate.

	\subsection{Comparison between our method and traditional GP on the \textsl{Linear Elasticity} benchmark}
	\label{sec:LEcomparison}
\begin{figure}[t!]%
		\centering
		\includegraphics[scale=1]{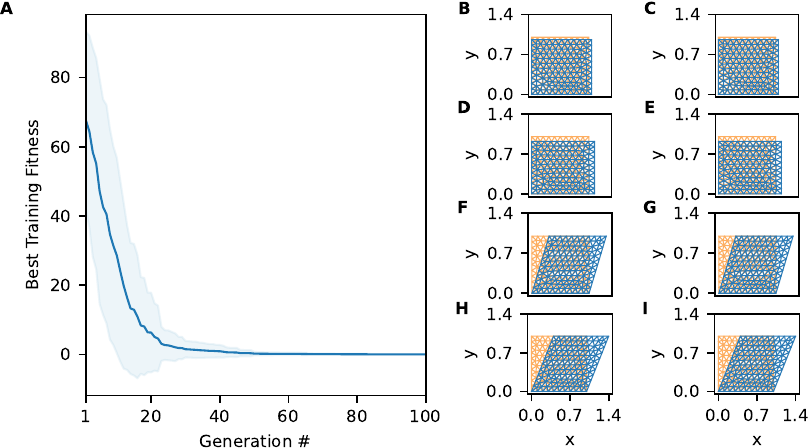}
		\caption{Result for the \textsl{Linear Elasticity} problem using traditional genetic programming. (A) Evolution of the mean and standard deviation of the absolute value of the best training fitness over the model discovery runs $(n=50)$. Comparison between the deformed configurations (B, D) for $\epsilon \in \{0.01, 0.02\}$ (uniaxial tension) and (F, H) for $\gamma \in \{0.06, 0.08\}$ (pure shear) corresponding to the best model at the end of the model discovery stage and the associated test data (C, E) and (G, I), respectively. The magnitude of the displacements has been scaled by a factor of $5$ for clarity. Orange meshes represent the reference configurations.}
		\label{fig:gp_linear_elasticity_plots}%
	\end{figure}

	As discussed in Section~\ref{sec:linel}, the dataset used for \textsl{Linear Elasticity} only consists of homogeneous deformations. Hence, a traditional Genetic Programming-based SR offers an alternative strategy to learn eq.~\eqref{eq:lin_ela_energy}, where the energy is a function that maps scalar variables $F_{11}, F_{12}, F_{21}, F_{22}$, which represent the homogeneous deformation gradient. We compared our method with such an approach, where we used addition, subtraction, multiplication, division, square and square root as primitives and did not enforce any strong typing. After choosing the best set of hyperparameters through a grid search, we performed the $50$ model discovery runs to assess the performance of the traditional approach (Figure~\ref{fig:gp_linear_elasticity_plots}). Notably, the recovery rate dramatically drops  to $4\%$, since in the traditional SR tensorial operations (such as, trace and symmetric part) must be learned as well from component representations.

	Also, the traditional SR approach cannot be applied without introducing additional operators (such as discrete integration/differentiation operations) whenever the deformations are non-homogeneous. Hence, our approach provides a distinct advantage over traditional SR in the case of field problems. To better elucidate this aspect, we studied a genuine field problem, where a body force induces non-homogeneous deformations. In this case, eq.~\eqref{eq:lin_ela_energy} becomes
	\begin{equation}
		\frac{1}{2} \langle \bm{S},\bm{E} \rangle + \langle \bm{u}, \bm{f}\rangle,
	\end{equation}
	where $\bm{u}$ and $\bm{f}$ are the primal vector-valued $0$-cochains representing the (dimensionless) displacements of the primal nodes and the  body force at each primal node, respectively.
	
	We generated $10$ data samples where $\bm{u}$ is quadratic (constant body force) and given by:
	\begin{align}
		\label{eq:nonhomog}
		\bm{u}(x_{i,1}, x_{i,2}) = - \frac{a}{10}x_{i,1}^2\hat{\bm{e}}_1 + a(x_{i,2}^2 - 1)\hat{\bm{e}}_2, \quad 	i \in \{0, \dots, n_0\},
	\end{align}
	where $a \in \{0.01, 0.02, \dots, 0.1\}$ is a dimensionless constant and $(x_{i,1},  x_{i,2})$ are the (normalized) Cartesian coordinates of the $i$-th node and $\hat{\bm{e}}_i$ the corresponding orthonormal basis vectors. The associated body forces $\bm{f}$ are computed by taking the negative of the (discrete) divergence of the stress tensor $\bm{S}$, with the deformations and stresses related by eqs.~\eqref{eq:defgrad}-\eqref{eq:stress}. The full dataset (training, validation and test) consists of the $10$ newly generated $(\bm{X}, \bm{f})$ pairs, where $\bm{X}$ is the matrix of the current positions of the nodes, and the pure tension samples (10) of Section~\ref{sec:linel}. 
	
	\begin{figure}[t!]%
		\centering
		\includegraphics[scale=1]{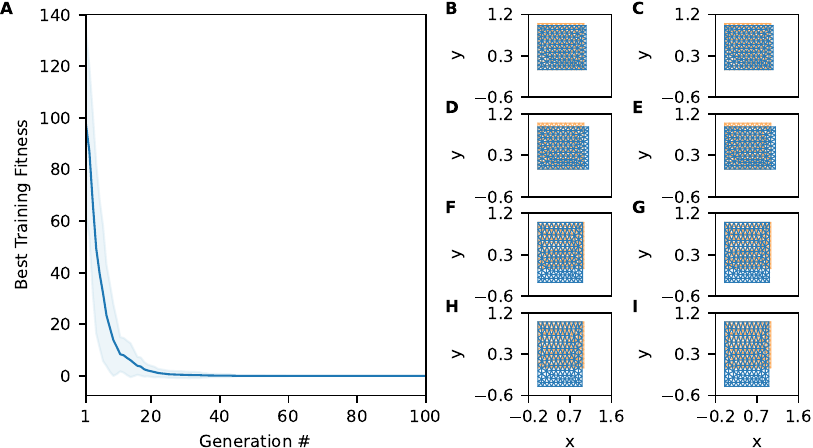}
		\caption{Result for the \textsl{Linear Elasticity} problem using a dataset containing non-homogeneous deformations. (A) Evolution of the mean and standard deviation of the absolute value of the best training fitness over the model discovery runs $(n=50)$. Comparison between the deformed configurations (B, D) for $\epsilon \in \{0.01, 0.02\}$ (uniaxial tension) and (F, H) for $a \in \{0.03, 0.04\}$ (non-homogeneous deformations, see eq.~\eqref{eq:nonhomog}) corresponding to the best model at the end of the model discovery stage and the associated test data (C, E) and (G, I), respectively. The magnitude of the displacements has been scaled by a factor of $5$ for clarity. Orange meshes represent the reference configurations.}
		\label{fig:linear_elasticity_non_hom_plots}%
	\end{figure}
	
	We performed $50$ model discovery runs using the best set of hyperparameters found for \textsl{Linear Elasticity}; the results are reported in Figure~\ref{fig:linear_elasticity_non_hom_plots}. The recovery rate dropped to $62\%$, probably because the hyperparameters were not optimal for the problem based on the new dataset.
	
\end{document}